\documentclass{IOS-Book-Article}

\usepackage{mathptmx}
\usepackage{soul}\setuldepth{article}
\usepackage{graphicx}
\usepackage{paralist}
\usepackage{comment}
\usepackage{placeins}

\usepackage{framed}

\def\hb{\hbox to 11.5 cm{}}
\begin{document}
%%%%%%%%%%%%%%%%%%%%%%%%%%%%%%%%%%%%%%%%%%%%%%%%%%%%%%%%%%%%%%%%%
\begin{frontmatter}

\title{Ontology Population using LLMs}

\markboth{}{August 2024\hb}
%\subtitle{Subtitle}

\author[A]{\fnms{Sanaz} \snm{Saki Norouzi}\orcid{0009-0003-4441-108X}%
\thanks{The first three named authors share first authorship.}},
\author[A]{\fnms{Adrita} \snm{Barua}\orcid{0000-0002-3287-7443}},
\author[B]{\fnms{Antrea} \snm{Christou}\orcid{0009-0003-0428-2439}},
\author[A]{\fnms{Nikita} \snm{Gautam}\orcid{0000-0002-1572-5405}},
\author[A]{\fnms{Andrew} \snm{Eells}\orcid{0000-0001-6357-6646}},
\author[A]{\fnms{Pascal} \snm{Hitzler}\orcid{0000-0001-6192-3472}},
\author[B]{\fnms{Cogan} \snm{Shimizu}\orcid{0000-0003-4283-8701}%
\thanks{Corresponding Author: Cogan Shimizu, \url{cogan.shimizu@wright.edu}}}

\address[A]{Kansas State University, Manhattan, Kansas, USA}
\address[B]{Wright State University, Dayton, Ohio, USA}

\runningauthor{Saki Norouzi, S., Barua, A., Christou, A. et al.}
%%%%%%%%%%%%%%%%%%%%%%%%%%%%%%%%%%%%%%%%%%%%%%%%%%%%%%%%%%%%%%%%%
\begin{abstract}
  Knowledge graphs (KGs) are increasingly utilized for data integration, representation, and visualization. While KG population is critical, it is often costly, especially when data must be extracted from unstructured text in natural language, which presents challenges, such as ambiguity and complex interpretations. Large Language Models (LLMs) offer promising capabilities for such tasks, excelling in natural language understanding and content generation. However, their tendency to ``hallucinate'' can produce inaccurate outputs. Despite these limitations, LLMs offer rapid and scalable processing of natural language data, and with prompt engineering and fine-tuning, they can approximate human-level performance in extracting and structuring data for KGs. This study investigates LLM effectiveness for the KG population, focusing on the Enslaved.org Hub Ontology. In this paper, we report that compared to the ground truth, LLM's can extract $\approx 90\%$ of triples, when provided a modular ontology as guidance in the prompts.
\end{abstract}
%%%%%%%%%%%%%%%%%%%%%%%%%%%%%%%%%%%%%%%%%%%%%%%%%%%%%%%%%%%%%%%%%
%%%%%%%%%%%%%%%%%%%%%%%%%%%%%%%%%%%%%%%%%%%%%%%%%%%%%%%%%%%%%%%%%
\end{frontmatter}
%%%%%%%%%%%%%%%%%%%%%%%%%%%%%%%%%%%%%%%%%%%%%%%%%%%%%%%%%%%%%%%%%
\markboth{September 2024\hb}{September 2024\hb}
%%%%%%%%%%%%%%%%%%%%%%%%%%%%%%%%%%%%%%%%%%%%%%%%%%%%%%%%%%%%%%%%%
%%%%%%%%%%%%%%%%%%%%%%%%%%%%%%%%%%%%%%%%%%%%%%%%%%%%%%%%%%%%%%%%%
\section{Introduction}
\label{sec:intro}
%%%%%
% Opening
% KGs are maturing, 
Knowledge graphs (KGs) have quickly become a major paradigm supported \cite{hitzler-cacm,kgs} by a broad set of methods and tools for the creation, extraction, integration, representation, and visualization of data, and supported by long-established W3C standards and recommendations \cite{owl2-primer,rdf-spec,shex,shacl}. 
% KE is expensive
Unfortunately, many aspects of knowledge engineering are quite expensive: from the knowledge model (i.e., ontology) development to the actual population (and validation) of the resulting KG \cite{comodide-eval}. Population is, in some sense perhaps the easiest component, yet this is only true if the data is already in a machine-parseable format (as many tools provide such a service, e.g., OpenRefine \cite{open-refine}). However, when the text is in natural language, this becomes problematic. Given the various complexities of interpreting sense, sentiment, frames, or anaphora, translating to facts or knowledge can be quite difficult. There are many natural language processing (NLP) techniques for tackling these problems, yet at the forefront -- in terms of both popularity and broad applicability -- are large language models (LLMs).

% LLMs are emerging
LLMs have rapidly emerged as powerful tools in various domains, showcasing remarkable proficiency in tasks such as natural language understanding, translation, and content generation \cite{llms,llms-attention}. 
% and they are surprisingly good at certain tasks
Their ability to process -- and interpret -- vast amounts of text data allows them to generate coherent and contextually relevant responses, often surpassing traditional models in creativity and nuance.

% But not always
However, LLMs are prone to an effect called confabulation, also referred to hallucination. This phenomenon occurs when the model produces responses that may \emph{seem} coherent and contextually relevant but are factually inaccurate or entirely fabricated. There is no strict guarantee that any particular response does not have confabulation, but they are more likely to occur in certain scenarios, such as when the query is ambiguous, the topic in question is particularly niche (and thus not present in the original training data), or if the query requires a certain level of complexity or creativity to accomplish.

% LLMs are cheaper than humans
Even with these caveats, LLMs are much faster than humans at certain knowledge extraction tasks (e.g., ingesting, translating, and extracting from natural language), and especially at volume.

% Yet they can be good for this particular task
With appropriate guidance (e.g., through prompt engineering, retrieval augmented generation \cite{lewis2020retrieval}, or fine-tuning \cite{llms-fine-tuning}) LLMs can approach human-level performance on such tasks.
% Therefore, explore if the Ontology Population task can be realistically done by LLM
Therefore, we wish to explore how effective LLMs can be in specifically populating a KG. In particular, we take this to mean, how well can an LLM extract or otherwise transform unstructured natural language into an output constrained by the ontology (i.e., the schema) for the KG.
% RQs
Succinctly, these research questions are as follows.\smallskip

\begin{compactenum}[\bfseries \qquad{RQ}1]
  \item Are LLMs capable of \emph{effective} KG population?
  \item What factors contribute to that effectiveness?
  \item Of the available LLMs, which perform best for the KG population task?
\end{compactenum}\smallskip

% Case Study: Enslaved.org Hub Ontology
To answer these questions, we have selected a KG where we are aware of multiple -- significantly different -- schemas and data sources for the same data. In particular, we have chosen the Enslaved.Org Hub Ontology, which seeks to (re)construct the narratives of historically enslaved peoples \cite{enslaved-jws} irrespective of their notability or infamy. Much of this data is manually curated from primary or secondary sources \cite{site:enslaved}. However, in many cases, there is overlap between public data sources (e.g., Wikipedia \cite{site:wikipedia} and Wikidata \cite{site:wikidata}). Due to the variety of structured and unstructured (i.e., natural language) methods for representing the same data, this provides an opportunity to reasonably assess the ability of LLMs to extract the data and evaluate whether or not that the extracted data is valid for the purposes of populating our manually curated ontology.

% Directory Info for rest of paper
The rest of this chapter is organized as follows.
% Case Study
Section~\ref{sec:data} introduces our case study: the Enslaved.Org Ontology and knowledge graph, as well as how we prepare the data for our experiments.
% Methodology
Section~\ref{sec:method} provides an overview of our methodology.
% Evaluation
In Section~\ref{sec:eval} we present our results and an evaluation thereof.
Finally, we discuss related work in Section~\ref{sec:related} before concluding in Section~\ref{sec:conc}.

%%%%%%%%%%%%%%%%%%%%%%%%%%%%%%%%%%%%%%%%%%%%%%%%%%%%%%%%%%%%%%%%%
\section{Background \& Case Study}
\label{sec:data}
%%%%%
In this section, we briefly present some preliminary background knowledge that will be useful for understanding the rest of the chapter. In particular, we present definitions for terms and notational and graphical conventions for our figures. We also provide a brief description of the case study that drives our experiment and evaluation.

%%%%%%%%%%%%%%%%%%%%%%%%%%%%%%%%%%%%%%%%%%%%%%%%%%%%%%%%%%%%%%%%%
\subsection{Schema Diagrams}
%%%%%
From the Modular Ontology Modeling (MOMo) methodology \cite{momo-swj}, which was used in the development of the Enslaved.org Hub ontology -- as appears in our case study, we use a (visual) graphical structure called a schema diagram as our primary method for communicating ontological structure. This diagram carries a reduced semantics; it is meant to be intuitive and easily understood, rather than explicitly and visually conveying the exact, underlying logical axioms. We use a consistent visual syntax across all diagrams.

Boxes of any non-gray color indicate a class. Goldenrod boxes are atomic classes. Purple boxes indicate a class that strictly consists of a controlled set of individuals. Blue boxes (with dashed borders) indicate \emph{hidden complexity} --- i.e., that there are additional relations, but which have been removed from view for clarity. Frequently, this means that it represents a class that is drawn from outside of the KWG namespaces (e.g., OWL Time \cite{time-tr}). Large gray boxes that encapsulate many arrows and boxes indicate a module, meaning that they are conceptually related. Yellow ellipses indicate a datatype. These are generally prefixed with the appropriate namespace, for clarity. Filled arrows indicate a binary relation. If one points to a box, then it is an object property. If one points to an ellipse, it is a data property. Open-face arrows indicate a subclass relation.

%%%%%%%%%%%%%%%%%%%%%%%%%%%%%%%%%%%%%%%%%%%%%%%%%%%%%%%%%%%%%%%%%
\subsection{Ontology Design Patterns \& Modules}
%%%%%
Ontology design patterns (ODPs) are tiny, self-contained ontologies that are focused on solving specific modeling problems \cite{odp1}. They are inspired by software engineering design patterns, and can be applied architecturally (i.e., a way to structure data) or exist on a spectrum from abstract to content-driven applications.
% Patterns to Modules
ODPs, especially through the MOMo methodology, are \emph{modularized} through a process called \emph{template-based instantiation} \cite{template}, which is then followed by \emph{systematic axiomatization} to produce a formalized model from the ODP using a series of 17 frequently used axioms for each node-edge-node construction in the schema diagram for the ODP or module \cite{owlaxax}. The use of patterns, by moving from top-level or abstract concepts and subsequently customizing for a specific (narrower) use-case is meant to mimic the human conceptualization process, which in some sense can be viewed as an analogy-driven process \cite{modont}.

% Ecosystem
By now, have a healthy ecosystem of various interactions, including libraries of ODPs (e.g., \cite{modl,csmodl}), an annotation language for describing pattern-based and modular structure \cite{opla,opla-cp}, and a tool ecosystem for creating pattern-based or modular ontologies \cite{comodide-eval,opla-tool}.

% Other Modular/pattern-based Knowledge Graphs and Ontologies
Indeed, modular ontologies have a young, but rich history. Examples span many years, from more bespoke project-focused ontologies (e.g., \cite{task-ont}) to the largest public geospatial KG in the world \cite{kwg-aimag,kwg-jws,kwg-fois}.
Most pertinent for this chapter, is the Enslave.org Hub's modular ontology, which we describe in the next section.

%%%%%%%%%%%%%%%%%%%%%%%%%%%%%%%%%%%%%%%%%%%%%%%%%%%%%%%%%%%%%%%%%
\subsection{The Enslaved.org Hub \& Ontology}
%%%%%
The scourge of African enslavement was fundamental to the making of Europe, Africa, the Americas, and Middle East and parts of the Asian subcontinent. The enduring legacies of black bondage shape the moral questions of humanity in our times. We have seen in the past decade a growth in interest in the subject in film, on television, and in historical fiction. Historians have spilled much ink writing monographs aimed primarily at other scholars. At the same time, however, it is a worthy goal to expand the production of scholarly output and to bring what historians do to the general public. 

Recently, there has been a significant shift in perceptions about what we can know about Enslaved Africans, their descendants, and those who asserted ownership over them throughout the world. As a result, a growing number of collections of scanned original manuscript documents, digitized material culture, and databases, that organize and make sense of records of enslavement, are free and readily accessible for scholarly and public consumption. Although this data is available through individual data silos, this proliferation of different projects and databases presents scholars, students, and the interested public with a number of challenges:
\begin{inparaenum}[\bfseries (a)]
  \item hyper-focused data for individual projects,
  \item little-to-no consistent disambiguation across projects,
  \item no data clearinghouse for this sort of data,
  \item query federation is not supported, 
  \item no established best practices, and
  \item lack of incentive to publish data.
\end{inparaenum}

The Enslaved.Org project introduces a new collaborative model for humanities scholarship by bringing together various professionals to share and expand knowledge about slavery \cite{enslaved-wapo,enslaved-acm}. This approach encourages scholars to rethink the way research is produced and shared, aiming to transform both academic practices and historical perspectives on slavery. The technical goal of Enslaved.Org established the Enslaved.Org Hub,a website that provides one-stop querying and inspection capabilities for integrated historic data on the slave trade, originating from a diverse set of data sources and contributors, thereby allowing students, researchers and the general public to search over numerous databases to understand and reconstruct the lives of individuals who were part of the historical slave trade \cite{site:enslaved}.

To address the underlying data integration issues, Enslaved.Org opted to follow the state of the art by establishing a KG equipped with an ontology as a schema. The Enslaved Ontology serves as the underlying schema and data organization paradigm for the Enslaved Hub. It is not used for reasoning or inference, but as a guide for organizing and integrating the data, and understanding the knowledge base as a whole. For additional detail on the modeling approach see \cite{enslaved-workflow}, and for a thorough examination of the ontology and its core concepts are detailed in \cite{enslaved-jws, enslaved-tr}.

%%%%%%%%%%%%%%%%%%%%%%%%%%%%%%%%%%%%%%%%%%%%%%%%%%%%%%%%%%%%%%%%%
\subsection{Wikidata \& Patterns}
%%%%%
Recently, \emph{community-driven KG plaforms} have grown in interest. These are software that hosts a KG that accepts community data from community sources (i.e., data that comes from outside the original development team) and, in general, have a focus on modeling and presenting provenance and lineage of the constituent data. With the growth of the use of KGs, some communities and larger constituencies are being left behind because many of the human-machine interfaces for KGs require more advance technical skills. To address this, the platform, Wikibase, can be deployed to mitigate issues of access for less technically practiced community members. Wikidata is the pre-eminent public Wikibase installation hosted by the Wikimedia Foundation \cite{site:wikidata}.

% Wikibase
Using the Wikibase platform has many advantages: an out of the box software for de-referencing, a convenient user interface for the less technically practiced, a consistent way to track and record provenance and lineage of data, and the option to execute SPARQL queries against an RDF representation of the knowledge graph. However, the provenance mechanism and the exact nature of the structure of the Wikibase representation can complicate developing a principled schema for KGs, as well as the approach to the materialization of the data for upload to the platform.

% Brief related work.
Indeed, other institutions, such as the European Union (EU), have also come to the conclusion that utilizing Wikibase can serve as the foundation for a community-driven knowledge graph. For instance, the EU Knowledge Graph\footnote{\url{https://linkedopendata.eu/wiki/The_EU_Knowledge_Graph}} \cite{DBLP:conf/semweb/DiefenbachWA21} is deployed on a Wikibase installation, as is the Disability Wiki that serves as an metadata knowledge graph for community documents \cite{DBLP:conf/i-semantics/BisenAMCGKMMDDG22}.

% Wikibase patterns
Recently, we have created a library of design patterns for Wikibase \cite{wikibase-patterns} that allow for facilitated transition from traditional ontology modeling, to the structure that would be expected so as to conform with Wikibase's underlying semantic structure.

%%%%%%%%%%%%%%%%%%%%%%%%%%%%%%%%%%%%%%%%%%%%%%%%%%%%%%%%%%%%%%%%%
\section{Knowledge Graph Population Methodology} 
\label{sec:method}
%%%%%
Our methodology consists of three stages:
\begin{inparaenum}[\bfseries (a)]
  \item Data Pre-processing,
  \item Text Retrieval, and
  \item KG Population.
\end{inparaenum}
This is followed by an evaluation, which we describe in Section~\ref{sec:eval}.

%%%%%%%%%%%%%%%%%%%%%%%%%%%%%%%%%%%%%%%%%%%%%%%%%%%%%%%%%%%%%%%%%
\subsection{Data Pre-processing}
%%%%%
This step consists of two sub-steps. First, the data needs to be collected, collated, and curated. Specifically, this means we needed to find relevant natural language text, such as from Wikipedia \cite{site:wikipedia}, that covered enslaved persons such that they could be found in Enslaved.org and Wikidata.

These articles needed to be cleaned for formatting. This included removal of headers and citations. Additionally the text taken from Wikipedia had the infobox removed, as exactly that information would be found in Wikidata.

%%%%%%%%%%%%%%%%%%%%%%%%%%%%%%%%%%%%%%%%%%%%%%%%%%%%%%%%%%%%%%%%%
\subsubsection{Dataset Collection, Collation, \& Curation}
%%%%%
This step essentially prepares quality data for our experiments. We took the following actions.
\begin{compactenum}
  \item First, a list of those individuals that exist on both Wikidata and Enslaved.org needed to be identified.
  \item Next, we used string matching between sources to help narrow possible matches before having to manually choose.
  \item Then, for each individual where we had data from both sources, we: individual:
  \begin{compactenum}
    \item downloaded the Wikitext from Wikidata\footnote{\url{https://wikidata.org/}}
    \item downloaded the TTL file from Enslaved.org\footnote{\url{https://lod.enslaved.org/}}
    \item read the TTL file into a graph using RDFlib\footnote{\url{https://rdflib.readthedocs.io/en/stable/}}
    \item all triples where individual was the subject were written to a tab-separated value (TSV) file
  \end{compactenum}
\end{compactenum}
As we knew that we would be eventually sourcing data from a natural language driven process (i.e., the LLM), we opted for a simpler output structure: the TSV. Specifically, this allowed us to have a minimal syntax mechanism by which we could compare two outputs, without worrying about small errors in syntax.

%%%%%%%%%%%%%%%%%%%%%%%%%%%%%%%%%%%%%%%%%%%%%%%%%%%%%%%%%%%%%%%%%
\subsubsection{Module Preparation}
%%%%%
For both of the two different schemas, we generated and filtered ontology modules based on schema relationships presented in both the Wikibase \cite{wikibase-patterns} and Enslaved \cite{momo-swj} ontologies.
We focused on extracting and utilizing module information and attributes relevant to the content available in the Wikipedia text files. Modules and attributes that did not provide useful information from the text files were skipped, ensuring that only the relevant data was included during the ontology population process. 

For example, in the Wikibase schema, we chose to skip the relation \texttt{isDirectly} \texttt{BasedOn(Agent\_Information, Reference)}, which typically provides information about the source from which the particular agent information was populated. Since we used Wikidata as the source to populate the ontology, this relation did not necessarily provide meaningful information about the actual source of the data, as it might not point to the original reference material. 
%%%%%
\begin{figure}
  \centering
  \includegraphics[width=0.5\linewidth]{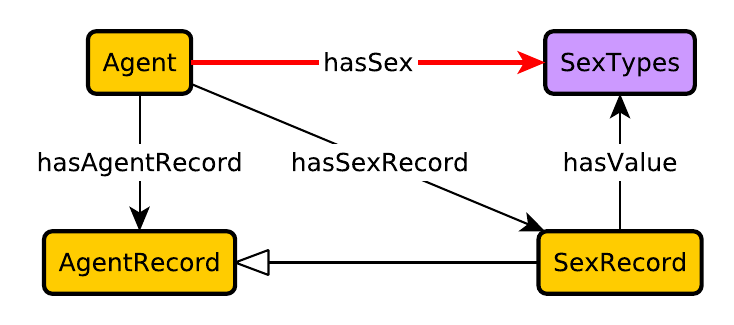}
  \caption{A graphical demonstration of how role chains in the Enslaved Ontology were ``collapsed'' into simpler shortcuts to generally improve data extraction performance.}
  \label{fig:prop-collapse}
\end{figure}
%%%%%
Similarly, for the Enslaved Ontology, we simplified chained relations to make them easier for the LLMs to capture. For instance, in the Sex Record Module, the original structure is \texttt{Agent} has \texttt{AgentRecord}, \texttt{SexRecord} is-a \texttt{AgentRecord}, and \texttt{SexRecord} \texttt{hasValue} \texttt{SexTypes}. We simplified this to a single relation: \texttt{hasSex(Agent,} \texttt{Sex\_Type)}, as shown in Figure~\ref{fig:prop-collapse}. Similar simplifications were made for other modules to enhance clarity and usability, ensuring the LLMs could effectively process and understand the relations. These shortcuts are included in our online documentation.
%
% \cognote{Additionally, we skipped relations that did not provide useful information in this context.}

%%%%%%%%%%%%%%%%%%%%%%%%%%%%%%%%%%%%%%%%%%%%%%%%%%%%%%%%%%%%%%%%%
\subsection{Relevant Text Retrieval}
%%%%%
Given the large size of the text files crawled from Wikipedia, which often exceeds the context window (and thus token limit) of LLMs, it was necessary to determine methods for passing only maximally relevant information to the LLM for populating the ontology. We used two methods for this purpose: text summarization and RAG, and we tested them both on GPT-4 and Llama-3 with temperature set to zero.

%%%%%%%%%%%%%%%%%%%%%%%%%%%%%%%%%%%%%%%%%%%%%%%%%%%%%%%%%%%%%%%%%
\subsubsection{Text Summarization}
%%%%%
\begin{figure}[tb]
  \includegraphics[width=\textwidth]{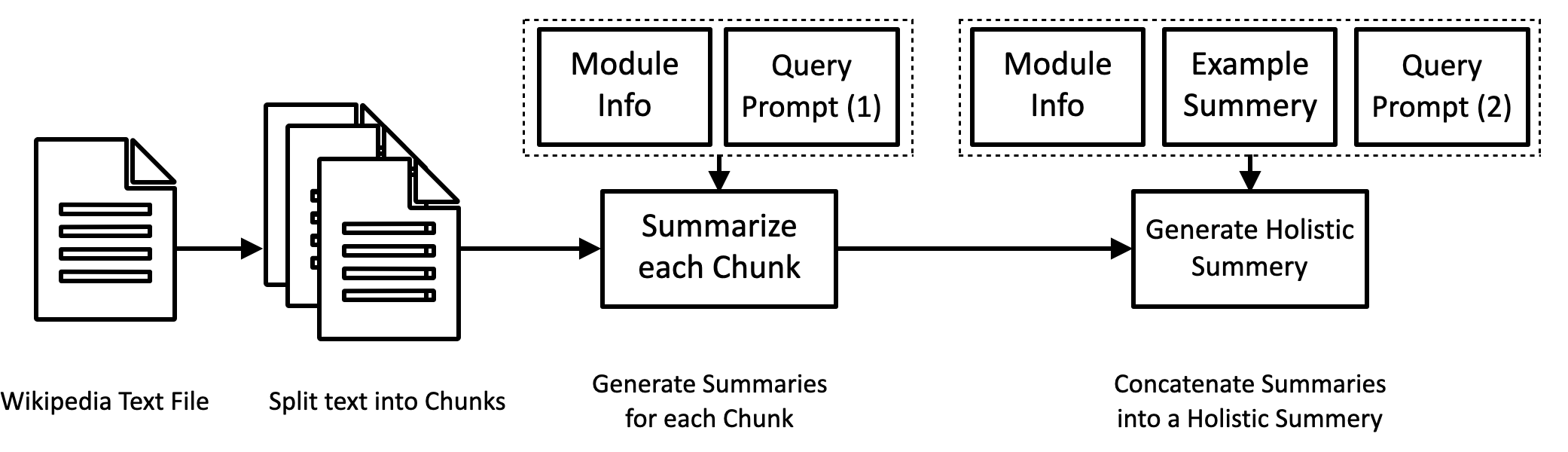}
  \caption{Summarizing the text files using few shot prompting}
  \label{fig:summarization}
\end{figure}
%%%%%
In the summarization method, LLMs are used to generate concise summaries of the relevant texts from the dataset. These summaries serve as a condensed form of information that can be used in the subsequent ontology population step. We employ a few-shot prompting technique with LLMs to generate summaries that are specifically relevant to the ontology modules, a task that traditional summarization tools cannot accomplish. Figure \ref{fig:summarization} shows the steps for the summarization process.

\noindent\textbf{Text Chunking:}
The text files are first split into manageable chunks, ensuring that each chunk stays within the model's token limit. This step involves reading the text file, splitting it into paragraphs, and then concatenating these paragraphs into chunks.

\noindent\textbf{Module Integration:}
A predefined module file containing relevant schema relationships was used to guide the summarization process. This file was read and incorporated into the prompts fed to the LLM, ensuring that the summaries generated were aligned with the desired ontology structure.

\noindent\textbf{Summary Generation:}
For each chunk, a query prompt was created that included both the text chunk and the relevant module content. The LLM model was then used to generate a summary that focused on extracting and retaining the most pertinent information based on the predefined modules. We provide our query prompt for the summarization.

\begin{figure}[!ht]
\raggedright
\footnotesize
\begin{framed}
Summarize the following text, by keeping the relevant information from these modules (if any): \{module\_content\}. 
The birth and death dates are mentioned in parenthesis after the agent name. For example : Joseph Vance Lewis (December 25, 1853 – April 24, 1925), was a slave who was freed. Here the Birth date is December 25, 1853 and Death date is April 24, 1925. So keep those information in the summary. The given text to summarize: \{text\_chunk\}
\end{framed}
\vspace{-0.5cm}
\end{figure}

\noindent\textbf{Holistic Summary Creation with Few-Shot Learning:}
After generating individual summaries for each chunk, these were concatenated to form a complete context. To ensure consistency and relevance, an example summary was provided as part of the prompt for holistic summary generation, employing few-shot learning. A final holistic summary was then generated, which synthesized the key points from all chunks, ensuring that the summary adhered to the structure and details outlined in the modules and the example summary. We also explicitly mention additional information about certain modules (e.g., the Interagent Relationship Record Module) to help guide the LLM in understanding the module contents in relation to the ontology, which may not be as straightforward in other cases (e.g., the Sex Record Module). We provide our query prompt for this process.

\begin{figure}[!h]
\raggedright
\footnotesize
\begin{framed}
Query Prompt: "Here is an example summary that highlights the key points from a text file based on the given modules. The text file discusses a single agent or person who is introduced initially. So, the summary should focus solely on that particular agent. The birth and death dates are mentioned in parenthesis after the agent name. For example : Joseph Vance Lewis (December 25, 1853 – April 24, 1925), was a slave who was freed. Here the Birth date is December 25, 1853 and Death date is April 24, 1925. The Interagent Relationship Record Module describes whether the agent has a relationship with another Enslaver or Owner. The Person Status Module indicates whether the agent is an enslaved person and mentions any status-generating events.

Example summary: \{example\_summary\}

Please provide a holistic summary from the given text that follows the format of the example summary and keeps the relevant information from these modules (if any): \{module\_content\}. 

The given text is: \{concatenation of summaries from each chunks\}"
\end{framed}
\end{figure}

%%%%%%%%%%%%%%%%%%%%%%%%%%%%%%%%%%%%%%%%%%%%%%%%%%%%%%%%%%%%%%%%%
\subsubsection{Retrieval-Augmented Generation}
%%%%%
Retrieval Augmented Generation (RAG) is a machine learning approach used in NLP that integrates a retrieval system to improve the performance of a generative model \cite{lewis2020retrieval}.

The core idea is to leverage the strengths of retrieval-based methods and generative models to produce more precise results. In this setup, the relevant information is retrieved based on the user’s query from an external knowledge source, such as a database or a knowledge graph. Thus, to implement the RAG mechanism, all extracted text files from Wikipedia are first divided into smaller chunks to ensure efficient processing and then embedded into a vector database. This chunking process helps manage large documents by breaking them into manageable sections, each of which can be individually compared for relevance. Two distinct prompt models are employed during the process: the first model explicitly instructs the system to populate the ontology for the agent, indicating that the document begins with a specific name and that the content is primarily about that entity. The second model, in contrast, omits this specific instruction, allowing for a more general approach to content retrieval.

In the RAG mechanism, the input prompt (query) is also transformed into an embedding, which is then compared against the pre-embedded document chunks. The similarity between the prompt embedding and the document chunk embeddings determines which chunks are most relevant. These selected chunks, based on their high similarity scores, are retrieved and used in conjunction with the original query as input to the language models. This approach enhances the system’s ability to generate informed and context-rich responses by integrating relevant document content directly into the model’s output generation.

For the implementation of the RAG mechanism, LangChain \footnote{https://www.langchain.com/} was employed as the framework to handle the retrieval and augmentation processes. The embeddings used for both the documents and queries are generated using Hugging Face's pre-trained model, specifically the sentence-transformers/all-MiniLM-L6-v2, which is designed to efficiently capture semantic similarities between text chunks. This combination of tools ensures accurate retrieval and seamless integration of relevant information into the language model’s output. Besides using GPT-4 (temperature set to zero), we use the open-source LLM ``Meta-Llama-3-8B-Instruct'' so as to ensure reproducibility, with the temperature set to zero to obtain deterministic results in this approach.

%%%%%%%%%%%%%%%%%%%%%%%%%%%%%%%%%%%%%%%%%%%%%%%%%%%%%%%%%%%%%%%%%
\subsection{Populating Ontology Modules}
\label{ssec:populating}
%%%%%
% Mention the prompts used in each step.
%%%%%
After extracting the relevant texts from each text file, we implemented a systematic approach to populate ontology modules using the extracted information. A predefined module file containing schema relationships guided the ontology population process. This module file was read and integrated into the system, serving as the foundation for structuring the ontology.

A detailed query prompt was then constructed, which included an example format demonstrating how the ontology should be populated based on the information available in the text files, utilizing a few-shot learning approach. The query prompt was explicitly designed, specifying the structure and terminology to be used for each ontology module.
For instance, the example format used in the prompt for Age record module would be,  
hasAgeValue(Agent, xsd:double): hasAgeValue(Absalom Jones, 71).

In general, ontology modules were populated according to the schema relationships for each module and attributes described for that relation in the text files. The prompt also included instructions to skip any relations for which no information was provided in the text files, ensuring that only relevant data was captured. 

The LLM was then prompted with the query, the content of the retrieved text files, and the schema modules. This combination allowed the model to accurately extract and structure the information according to the predefined ontology format, ensuring consistency and alignment with the specified schema. Both the Llama-3 and GPT-4 models were employed with a temperature setting of 0 for the prompts to populate the ontologies. 

%%%%%%%%%%%%%%%%%%%%%%%%%%%%%%%%%%%%%%%%%%%%%%%%%%%%%%%%%%%%%%%%%
\section{Evaluation}
\label{sec:eval}
%%%%%

%%%%%%%%%%%%%%%%%%%%%%%%%%%%%%%%%%%%%%%%%%%%%%%%%%%%%%%%%%%%%%%%%
\subsection{Dataset used}
%%%%%
We aim to evaluate the triple generation accuracy of a Large Language Model (LLM) against reference data from Tab-Separated Values (TSV) files. These TSV files are our reference data and include information in triple format ( subject, predicate, value) about every individual from the Enslaved ontology.

% Previous pipeline:
%In order to perform this evaluation, we compare the features and variables from the TSV files with those from the TXT files, which were produced using various processes. The target data, or the output we want to match the reference data with, is represented by the TXT files. We compare these datasets using several string similarity metrics to see how closely the generated triples from the LLM match the data in the TSV files.

% New pipeline:

We organize the data for simpler comparison by first converting the generated text (TXT) files into tab-separated values (TSV) files in order to evaluate the language model's (LLM) performance. The quality of the data produced can be evaluated by comparing it to the truth data that is already in TSV format after the text of the TXT files has been parsed to extract relevant triples. By using a variety of string similarity variables, this comparison allows us to measure the degree to which the produced triples match the reference data.
%%%%%%%%%%%%%%%%%%%%%%%%%%%%%%%%%%%%%%%%%%%%%%%%%%%%%%%%%%%%%%%%%
\subsection{Similarity Metrics}
%%%%%
To make sure that both typographical errors and semantic differences are properly taken into account when assessing the alignment of characteristics and values between our generated files, we conduct a reconciliation step using a number of string similarity measures identified using guidance from \cite{10.1007/978-3-642-41338-4_19}. 

\noindent\textbf{Cosine Similarity} -- which calculates the cosine of the angle between two term frequency-inverse document frequency (TF-IDF) vectors -- becomes useful \cite{10.1007/978-3-642-41338-4_19}. It should be noted that cosine similarity can fail to accurately represent the subtle differences in very short texts or texts with similar structure but different contents. Short texts frequently produce less informative similarity scores and less discriminative TF-IDF vectors.

\noindent\textbf{Fuzzy Matching}: Creates similarity ratios that take into consideration possible typos and incomplete matches by using the \texttt{rapidfuzz} package \cite{Bosker2021}. These two ratios are employed:
\begin{compactitem}
  \item \textbf{FuzzyWuzzy Ratio:} compares two strings' overall similarity.
  \item \textbf{FuzzyWuzzy Token Set Ratio:} allows for the comparison of strings that have different word ordering by taking into account the token sets of two strings.
\end{compactitem}

\noindent\textbf{Jaro-Winkler Similarity:} Calculates the similarity between strings by taking into account shared prefixes and small typographical changes, for the purpose of aligning attributes or values with comparable roots \cite{10.1007/978-3-642-41338-4_19}.

% Might not have to be included
%\noindent\textbf{BERTScore}: This computes a more sophisticated similarity score that takes into consideration semantic similarities between the predicates and values in the files by utilizing the BERT model \cite{zhang2020bertscoreevaluatingtextgeneration}. Because \textbf{BERTScore} can be computed for both predicates and values, it offers a reliable way to evaluate alignment accuracy.

%%%%%%%%%%%%%%%%%%%%%%%%%%%%%%%%%%%%%%%%%%%%%%%%%%%%%%%%%%%%%%%%%
\subsection{Evaluation Methodology}
\label{ssec:eval-meth}

From the original text (TXT) files, we first focus on creating tab-separated values (TSV) files to start the evaluation process. The purpose of this transformation is to organize the data in a way that makes comparison and analysis easier. The output produced by the language model (LLM) is contained in the TXT files, which are our target data—the values we want to match the reference data with. This is to essentially make the evaluation process more accurate since now we will compare pair of files with similar structure. 

The relevant triples, which usually consist of predicates and objects that point to the relationships within the data, are first extracted by parsing the text of each TXT file. We arrange the triples in an organized way when they have been detected, making sure that a tab character separates every element.   We can preserve data readability with this setup.

In order to compare the truth data with the generated triples, pairs of files—one storing the reference data and the other containing the language model (LLM) output—are evaluated. To evaluate the degree of similarity between related entries in the truth and created datasets, we use a variety of string similarity measures for each pair, including fuzzy matching and cosine similarity. Iteratively going over each triple in the truth file and comparing it to the matching triple in the produced file, scores are determined based on how closely they match.

After that, we define a set of similarity thresholds to determine whether triples that are produced match the truth data sufficiently.   Checking to see if any of the similarity metrics for both predicates and objects—exceed these predetermined thresholds is part of the evaluation process. Identifying unique truth predicates and objects enables us to calculate the total number of unique comparisons. The percentage of good matches in comparison to the total number of unique triples is subsequently calculated by counting the unique good matches that reach or exceed the specified thresholds. 

The performance of our produced outputs was then assessed by determining the average, minimum, and maximum percentages of good matches for each llm-output comparison. This analysis helped us assess the overall quality of our results by revealing how well the created triples aligned with the reference data. 

Furthermore, from every file and model, we extracted the highest values of the similarity metric.Understanding the top-performing data allowed us to identify the areas in which our models performed really well while also looking into weaknesses.

%%%%%%%%%%%%%%%%%%%%%%%%%%%%%%%%%%%%%%%%%%%%%%%%%%%%%%%%%%%%%%%%%
\section{Results and Discussion}
\label{sec:results}
%%%%%
\begin{table}[t]
\centering
\begin{tabular}{l|c|c|c|c|c}
    \textbf{LLM Model}  & \textbf{Avg \%} & \textbf{Ttl \%} & \textbf{\#F} & \textbf{Avg$_A$ \%} & \textbf{Ttl$_A$ \%} \\\hline
    GPT4\_Enslaved\_MainAgent & 82.30 & 81.60 & 14 & 88.55 & 88.09\\\hline
    GPT4\_Enslaved\_notrestrictedToMAgent & 87.87 & 86.82 & 4 & 89.11 & 88.69\\\hline
    GPT4\_WB\_MainAgent & 77.08 & 76.10 & 25 & 88.02 & 87.63\\\hline
    GPT4\_WB\_notrestrictedToMAgent & 85.03 & 84.12 & 7 & 87.69 & 87.34\\\hline
    GPT-4\_Summarization\_Enslaved & 81.16 & 80.86 & 0 & 81.16 & 80.86\\\hline
    GPT-4\_Summarization\_WB & 82.60 & 82.03 & 0 & 82.60 & 82.03\\\hline
    llama\_WB\_MainAgent & 71.17 & 70.88 & 6 & 73.64 & 73.20\\\hline 
    llama\_WB\_notrestrictedToMAgent & 71.25 & 70.63 & 2 & 71.92 & 71.41
\end{tabular}
\caption{Similarity matching between triples extracted by the LLM models plus prompting strategy, reporting the average and total coverage for each module and in aggregate (columns with subscript $A$). The \#F column indicates the number of modules for which triple extraction entirely failed (i.e., a coverage of 0).}
\label{tab:metrics_llm_models}
\end{table}
%%%%%%
% Description of how results were calculated
The LLM evaluation results are displayed in Table \ref{tab:metrics_llm_models}. For each LLM model and prompting strategy, extraction strategy, and source data, we evaluate how well the extracted triples match our ground truth. This is reported as \emph{coverage} per module and overall. A triple is considered to be covered if an extracted triple has a similarity 80\% in any of the metrics reported in Section~\ref{sec:method}.\footnote{It is generally considered to be a ``good'' match if a similarity metric returns 70\% or better. We tend to a stricter threshold.} This covers small typos in the text, but also small variants in naming. The most common example is the inclusion of a middle initial, but the rest of the extracted triple matches the ground truth.

% Description of Columns
Each LLM model had at least one module with 100\% coverage, except for \textsf{llama\_WB\_notrestrictedToMAgent}, which had a only a maximum of 90.1\%. In this case, 100\% indicates that all triples had an appropriate match. However, each experiment that did not use the summarization technique had modules where the model failed to extract any relevant triples, the LLM model refused to complete the task, or for some other inscrutable reason did not provide a list of triples appropriate for extraction and mapping. 

Overall, we see the best performance from GPT-4, basically irrespective of the prompting strategy utilized.

For example, \textsf{GPT-4\_Summarization\_Enslaved} has the lowest average and total coverage, and performs better than either of the experiments conducted with \textsf{llama} models. Interestingly, extraction of triples according to the Wikibase model had highest fail rate, but still overall had higher coverage than the \textsf{Llama} models.

Generally, we see that the extraction of triples according to a schema is generally effective, approaching 90\% coverage when the model performs the task.

%%%%%%%%%%%%%%%%%%%%%%%%%%%%%%%%%%%%%%%%%%%%%%%%%%%%%%%%%%%%%%%%%
\section{Related Work}
\label{sec:related}
%%%%%
We have selected the below related research since it is consistent with our goal of automating ontology engineering tasks by utilizing large language models and solving issues like inaccurate development of concepts. The research articles brought light on the current issues surrounding ontology building, particularly those related to scalability and domain-specific accuracy. These publications also point to limitations in ontology population methods that we hope to fill with more accurate and organized data extraction techniques.

%%%%%%%%%%%%%%%%%%%%%%%%%%%%%%%%%%%%%%%%%%%%%%%%%%%%%%%%%%%%%%%%%
\subsection{Ontology Engineering and Construction}
\label{ssec:oe}
%%%%%
\paragraph{Ontology Engineering with Large Language Models (2023):}
Large language models (LLMs) are investigated here as a potential tool for automating ontology engineering activities including relationship detection and entity extraction. Although LLMs have the potential to reduce manual tasks, they frequently provide concepts that are inaccurate or too broad, which can lead to errors in complicated domains \cite{ontologyengineering2023}.

We overcome these weaknesses in our approach by using module-guided summarization and few-shot prompting to make sure that the output is properly designed to meet the requirements and structure of the ontology. We give the model appropriate examples and schema-driven prompts, which guide it to focus on important relationships and entities, in contrast to general LLM approaches that might produce ambiguous notions. We reduce the possibility of excessive generalization and unimportant information through incorporating the predefined modules into the summary process and making sure that it is in line with the ontology's schema. This makes the results that are produced more accurate and appropriate for demanding ontology tasks.

\paragraph{Towards Ontology Construction with Language Models (2023):}
The goal of this work is to expedite the process of ontology generation by exploring the use of language models to automatically generate ontologies from text. Although the method works well, it has trouble guaranteeing that the generated ontologies are accurate and complete as well as catching details unique to a given domain \cite{funk2023ontologyconstructionlanguagemodels}.

On the other hand, our method focuses on carefully populating particular areas of the ontology by extracting relevant information and providing module-guided summaries, instead of automatically creating full ontologies. We maintain the accuracy and contextual relevance of the populated ontology by including the ontology schema into the summarization process and using structured prompts to target relevant content. This approach provides domain-specific relevance and minimizes errors in difficult domains by prioritizing exact text retrieval and alignment with the predetermined structure, so avoiding the frequent issues associated with broad or partial ontology development.

\paragraph{OLAF: An Ontology Learning Applied Framework (2023):}
With a focus on collecting and organizing information from massive datasets, OLAF presents a system for automating ontology learning. Although the system has its advantages, its scalability issues and potential for poor performance when dealing with noisy or unstructured data restrict its use in a variety of real-world situations \cite{SCHAEFFER20232106}.

By combining retrieval-based techniques with text chunking, our strategy, on the other hand, directly addresses the problem of large-scale data and the potential noisiness that might occur. To make sure the LLM uses important, organized content, we divide large data sets into manageable portions that fit within the token constraints of language models. Retrieval-Augmented Generation (RAG) and summarization are two other techniques that let us restrict our focus to the most important data. Targeted retrieval handles capacity concerns that systems like OLAF have by guaranteeing that, even when processing massive or unstructured data, the content used for ontology population remains accurate, relevant, and in line with the particular ontology structure.

%%%%%%%%%%%%%%%%%%%%%%%%%%%%%%%%%%%%%%%%%%%%%%%%%%%%%%%%%%%%%%%%%
\subsection{Ontology Evaluation and Alignment}
\label{ssec:oe-oa}
%%%%%
\paragraph{Automated Ontology Evaluation: Evaluating Coverage and Correctness using a Domain Corpus (2023):}
Using domain-specific databases, this research suggests an automated approach to assess the reliability and scope of ontologies. The method may not fully capture contextual significance or domain-specific details, and it is constrained by the quality of the input data, although offering a robust review \cite{10.1145/3543873.3587617}.

Instead, in order to guarantee accuracy and relevance, our method concentrates on the ontology population. We prioritize domain-specific details to be directly included into the ontology during development by utilizing module-guided summarization and few-shot prompting. Our methodology involves domain-specific details during the population phase, resulting in a more contextually correct and relevant ontology, whereas this paper's method focuses on evaluating an ontology's scope and trustworthiness after it has been produced, frequently missing key details.

\paragraph{String Similarity Metrics for Ontology Alignment (2022):}
The paper addresses ontology alignment using string similarity metrics, which is essential for connecting different ontologies. These measures increase alignment accuracy, but their use in tricky alignments is limited since they may have difficulty with semantic mismatches and lack taking context for words into account \cite{10.1007/978-3-642-41338-4_19}.

%%%%%%%%%%%%%%%%%%%%%%%%%%%%%%%%%%%%%%%%%%%%%%%%%%%%%%%%%%%%%%%%%
\subsection{Knowledge Graphs and Language Models}
\label{ssec:kgs-lms}
%%%%%%
\paragraph{Large Language Models and Knowledge Graphs: Opportunities and Challenges (2023):}
With the goal of enhancing knowledge extraction and reasoning, this research investigates possibilities for collaboration between large language models and knowledge graphs. It does, however, draw attention to issues with scalability, model interpretability, and the requirement for high-quality data in order to avoid mistakes in knowledge graphs \cite{pan2023largelanguagemodelsknowledge}.

We address data quality and interpretability issues early by organizing the ontology population process with precise, domain-specific summaries, guaranteeing that the ontology is provided with relevant and accurate content right from the start.

\paragraph{TEXT2KGBENCH: A Benchmark for Ontology-Driven Knowledge Graph Generation from Text (2023):}
TEXT2KGBENCH presents a benchmark for assessing ontology-driven methods for text-to-knowledge graph generation. Though its reliance on particular ontologies restricts its applicability to other domains and methods of knowledge representation, the benchmark standardizes evaluation measures \cite{mihindukulasooriya2023text2kgbenchbenchmarkontologydrivenknowledge}.

RAG is useful in our situation because it facilitates the extraction of relevant information from huge data sets, which is subsequently used to produce content that is directly related to the ontology. This method minimizes problems associated with large-scale data processing and maintains high-quality information by ensuring that only the most pertinent data is used.

%%%%%%%%%%%%%%%%%%%%%%%%%%%%%%%%%%%%%%%%%%%%%%%%%%%%%%%%%%%%%%%%%
\subsection{Knowledge Extraction and Prompt Engineering}
\label{ssec:extraction}
%%%%%
\paragraph{SPIRES: Structured Prompt Interrogation and Recursive Extraction of Semantics (2023):}
Using structured prompts for recursive semantic extraction, SPIRES offers a zero-shot learning approach for knowledge base population. The method works well, but it is constrained by the prompt design's quality and the model's comprehension of complicated or difficult ideas \cite{caufield2023structuredpromptinterrogationrecursive}.

\paragraph{Testing Prompt Engineering Methods for Knowledge Extraction from Text (2023):}
This research uses language models to assess different prompt engineering techniques for knowledge extraction from text. The study highlights the benefit of different prompts but also points out their inconsistent aspects and the need for domain knowledge in order to create well-designed queries \cite{polat2024testing}.

%%%%%%%%%%%%%%%%%%%%%%%%%%%%%%%%%%%%%%%%%%%%%%%%%%%%%%%%%%%%%%%%%
\subsection{Evaluation Metrics for Text and Knowledge Extraction}
\label{ssec:eval-metrics}
%%%%%
\paragraph{BERTScore: Evaluating Text Generation with BERT (2019):}
A more sophisticated evaluation than previous techniques, BERTScore presents a metric for text output quality using BERT embeddings. However, the measure is susceptible to the selection of pre-trained BERT models and might not completely encompass semantic variations in produced text, particularly in synthetic or non-literal settings \cite{zhang2020bertscoreevaluatingtextgeneration}.

%%%%%%%%%%%%%%%%%%%%%%%%%%%%%%%%%%%%%%%%%%%%%%%%%%%%%%%%%%%%%%%%%
\section{Conclusion}
\label{sec:conc}
%%%%%
Knowledge graphs, are by now, utilized across many domains and across enterprise and academia \cite{hitzler-cacm}. They are supported by a healthy and mature ecosystem of W3C standards \cite{owl2-primer,rdf-spec,shex,shacl} and tool ecosystems (e.g., Protégé or GraphDB). They excel at the integration of heterogeneous datasets and perspective by allowing for semantic harmonization. Yet, when encountering new data, mapping into an existing ontology can be quite expensive. This might be due to the sheer volume of data to be interpreted.

In this chapter, we have presented a mechanism for examining the efficacy of utilizing large language models to do the semi-automated population of a modular ontology. This pipeline uses both LLM-powered text summarization and retrieval-augmented generation processes for conducting ontology-guided information distillation from natural language into forms more amenable to limited context windows and translation into triples to be loaded into a KG.

In our evaluation, we have shown that extraction of appropriate triples, as guided by a schema (and specifically in the case of our experiments: a modular ontology) is effective, approaching 90\% coverage when the LLM performs the task. Performance is similarly high agnostic to the prompting strategy or mechanism to fit the text into the models' context windows. As such, we conclude that LLMs are a sufficient tool for the extraction of triples from text when guided by a (modular) schema via prompting.

\textbf{Future Work.} These experiments are preliminary, insofar that they demonstrate feasibility. The evaluation is comparatively shallow, demonstrating coverage in a naive way. Since we are comparing text to ground truth, which was manually curated not necessarily from the text at hand (e.g., other sources), it is not known if failure to extract a triple at a high enough similarity is due to the fact that that information was missing in the first place, or if the LLM somehow missed a particular fact.

As such, further experiments must be conducted. We envision some next steps to answer the following questions.
\begin{compactitem}
    \item Does this LLM triple extraction process perform better than a human? We know that it is certainly \emph{faster}, but does the increase in speed outweigh missing or incorrect facts?
    \item What graph structures (i.e., data organized in which ontological formalisms) are most easily extracted to from textual data? Experiments demonstrate that the modular ontology originally developed for The Enslaved.org Hub performs better, but what exact characteristics facilitate that increase in coverage?
\end{compactitem}

%%%%%%%%%%%%%%%%%%%%%%%%%%%%%%%%%%%%%%%%%%%%%%%%%%%%%%%%%%%%%%%%%
\paragraph*{Acknowledgements}
The authors acknowledge partial funding under National Science Foundation grants 2333532 and 2333782, and from Kansas State University's Game Changing Research Initiative (GRIP) program.

%%%%%%%%%%%%%%%%%%%%%%%%%%%%%%%%%%%%%%%%%%%%%%%%%%%%%%%%%%%%%%%%%
\bibliographystyle{vancouver}
\bibliography{refs}
%%%%%%%%%%%%%%%%%%%%%%%%%%%%%%%%%%%%%%%%%%%%%%%%%%%%%%%%%%%%%%%%%
\end{document}